# We Tweet Like We Talk and Other Interesting Observations: An Analysis of Modalities of English Communication

Josiah Paul Zayner

*Josiah.P.Zayner@NASA.gov*

## ABSTRACT

Modalities of communication for human beings are gradually increasing in number with the advent of new forms of technology. Many human beings can readily transition between these different forms of communication with little or no effort, which brings about the question: How similar are these different communication modalities? To understand technology's influence on English communication, four different corpora were analyzed and compared: Writing from Books using the 1-grams database from the Google Books project, Twitter, IRC Chat, and transcribed Talking. Multi-word confusion matrices revealed that Talking has the most similarity when compared to the other modes of communication, while 1-grams were the least similar form of communication analyzed. Based on the analysis of word usage, word usage frequency distributions, and word class usage, among other things, Talking is also the most similar to Twitter and IRC Chat. This suggests that communicating using Twitter and IRC Chat evolved from Talking rather than Writing. When we communicate online, even though we are writing, we do not Tweet or Chat how we write books; we Tweet and Chat how we Speak. Nonfiction and Fiction writing were clearly differentiable from our analysis with Twitter and Chat being much more similar to Fiction than Nonfiction writing. These hypotheses were then tested using author and journalists Cory Doctorow. Mr. Doctorow's Writing, Twitter usage, and Talking were all found to have very similar vocabulary usage patterns as the amalgamized populations, as long as the writing was Fiction. However, Mr. Doctorow's Nonfiction writing is different from 1-grams and other collected Nonfiction writings. This data could perhaps be used to create more entertaining works of Nonfiction.

## INTRODUCTION

The evolution of new dialects of spoken languages has often been studied and modeled with respect to the contribution of idiolects (1) and the impact of spatial pressures (2). With technology allowing ease of communication between spatially distant individuals, there is the possibility that modern dialects may form between members of social networks without similar geographical locations. This raises a somewhat semantic question: What constitutes a dialect? It is clear that graphemic communication can be very different from phonemic communication, as these have been suggested to be different dialects in English (3). However, overall this question is complicated and overlooks a simpler question: How similar or different are communication corpora from each other?

In the past 25-30 years modes of online communication, such as Chat, appear as a combination of phonemic and graphemic dialects, but it is unclear how much these modes differ. Modern modalities of communication were developed on the basis of previous forms of communication, but it is unclear if the language usage is more similar to written or spoken "dialects". In this study, I have analyzed the vocabulary from four different bodies of work: books from the Google Books project (4), chat logs from IRC, transcribed verbal communication (i.e., Talking), and over one million unique messages sent on Twitter. What I find is unexpected. Though Chat and Twitter could be considered forms of written communication, the word usage patterns are more similar to Talking. Comparisons between Fiction and Nonfiction writing show that Fiction more closely resembles Chat and Twitter. Using a case study of Author, Speaker and Twitter user Cory Doctorow I find that the way someone speaks can be as similar to Nonfiction writing as it is to Fiction writing but we incorrigibly Tweet like we speak.

**METHODS**

Analysis and fitting was done with custom Perl code and the [SciDavis](SciDavis) program. Capitalization was removed from all corpora for analysis.

*Google 1-grams*
The Google N-gram project has sought to create a comprehensive database of word usage in written books. A direct statistical interpretation of the corpus should be approached with caution. The corpus digitizes all symbolic representations in a book, including numbers, letters, and symbols. For example, "70%" is considered a gram. Also, Google recognizes on their website that the level of optical character recognition (OCR) is not perfect. These errors can make it difficult to define the probability of occurrence for N-grams because the same word could be inappropriately identified as two distinct words. I used the 2009 corpus as this project was started before the release of the 2012 corpus. Any non-letter based gram was parsed out, so that any gram with numbers, punctuation, or non-Arabic characters was removed. Any word that contained an occurrence of 3 or more letters in a row was also removed. I don't think there is any proper word in the English language that has this (ex. [https://books.google.com/ngrams/graph?content=aaaaaaa&year_start=1800&year_end=2000&corpus=15&smoothing=3&share=&direct_url=t1%3B%2Caaaaaaa%3B%2Cc0](https://books.google.com/ngrams/graph?content=aaaaaaa&year_start=1800&year_end=2000&corpus=15&smoothing=3&share=&direct_url=t1%3B%2Caaaaaaa%3B%2Cc0)) The corpus was still massive with over ~271 billion counts and ~3.7 million unique words.

*Twitter*
For the Twitter corpus, the live English Twitter data stream was searched for a space character " ". This prevented any biasing of the datastream besides biasing it for tweets with spaces. Tweets that contained only links to websites were removed, and only unique Tweets were included in the analysis. The resulting Twitter corpus contains over 1.25 million unique tweets and over 12 million counts and ~3 million unique words.

*IRC Chat logs*
The IRC chats logs are for two different chat channels. In IRC chat when people are responding specifically to another person they start the sentence with "<name> :" I removed these to prevent erroneous results showing names as highly used. This has around 9 million counts and ~155,000 unique words.

*Talking or Spoken Word (Santa Barbara Corpus of Spoken American English)*
This corpus from SBCSAE contains many different types of human conversation. It is the smallest corpus with only around 250,000 counts and around 11,000 unique words.

I don't mean to imply at all that my corpora are perfect. There are errors initially in them and errors from difficult parsing. What I will say is that I tried hard to make the corpora as error free as necessary.

**Cory Doctorow Corpora**
*Talking*
This data was taken from transcribed speaking engagements and Q&A sessions by Mr. Doctorow, obtained from the Youtube video service. Cory Doctorow is an author through multiple mediums, which made him an ideal candidate. This corpus contains 22,962 counts and 3,794 unique words.

*Twitter*
This constitutes ~500 tweets from Mr. Doctorow as that is all that could be obtained from Twitter after parsing out those that just contained links. This corpus contains 2,990 total counts and 1,383 unique words.

*Fiction*
This corpus contains the freely available works *Down and Out in the Magic Kingdom* and *Little Brother*. This compromises 237,958 total counts and 12,118 unique words.

*Nonfiction*
Contains works from Mr. Doctorow's Nonfiction news, blog, and opinion writing. It contains 22,187 counts and 4,757 unique words.

**RESULTS & DISCUSSION**

To estimate the distribution of word frequencies, histograms were created by randomly sampling corpora. From each corpora, >90% of the words fall in the smallest frequency count bin unless I make the bin size ridiculously small (i.e. thousands of times or millions of times smaller than the word with the most counts) (**data not shown**). This determined that most of the data in the corpora are infrequently used words. As an example, of the 3,769,991 unique words pulled from Google 1-grams, *the* is used with the highest frequency at 20,510,449,496 times (~8% of the corpus). While the 10,000th most frequently used word *handles* is used 1,813,662 times (~ $7\times10^{-6}$ % of the corpus) and the 100,000th word *appropriators* is only used 33,001 times(~$1\times10^{-7}$ % of the corpus).

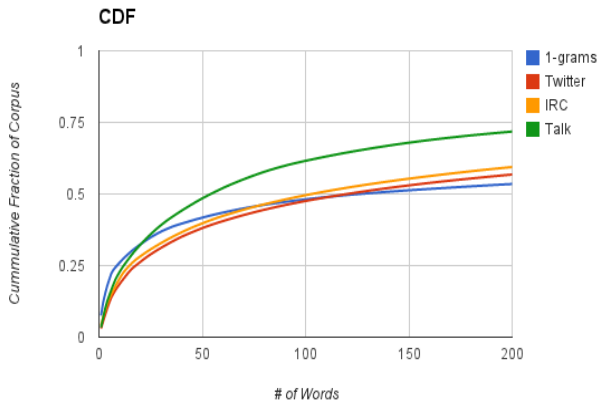

*Figure 1.* Cummulative Distribution Function plotted for the first 200 words of each corpus

Of the 3.7 million words most of them are used approximately a billion times less than the most used word. To analyze these corpora I needed to develop a normalization method. As a corpus becomes larger a single word losses importance because most of the words in a corpora are low occurrence. This can make some words relatively diluted. The goal of this study is to compare corpora to each other but using whole corpora would give erroneous results initially because the corpora are not all of equal size.

*Normalization*

To quantitatively describe the composition of each corpus, the Cumulative Distribution functions (CDFs) were plotted (**Fig. 1**). The CDFs show that >50% of the total word counts in each corpus are in the 150 highest ranked words or less. For further analysis, this was expanded to constitute the first 200 most popular words in each corpus. I fit the CDFs to double exponentials to look at the rate of CDF change per word (**Table 1** & **Fig. 1**). I did double exponential fits because single exponentials were disastrous due to drastic changes in corpus size growth per word over the length of the corpus. The 1-grams population has a faster growth rate in the beginning than every other corpora. About 20% of the 1-grams corpus is contained in the first 5 words. However, for Twitter 20% of the corpus is not reached until the 13th word, 11th for IRC Chat and 8th for Talking (**Table 1**). The second rate (Rate 2) however doesn't seem to make much intuitive sense as compared to the graph. The initial rates from Talk and 1-grams should be the highest and that is represented in Rate 1 clearly (**Fig. 1** & **Table 1**). However, the graph shows that Rate 2 is the greatest for 1-grams. Looking at the graph it should be the slowest as it has the lowest value at around 200 words. Maybe the program I used for fitting, SciDavis, is just whack but I don't think so. It has been very good to me in the past. Instead, I did a linear fit of the final slope of the Final 100 and Final 150 values of the CDF (**Table 1**). What is seen is much more intuitive and reasonable. Talking has the largest slope/fastest growth for both the Final 100 and the Final 150 words. This makes complete sense as the other data sets reach 50% of the cumulative fraction of the corpus by 100-130th word while Talking reaches it at the 55th. This final slope seems to be strongly dependent on total corpus size. Logically this makes sense. For example, If there is a corpus with 10 words with high occurrence and 1 quadrillion words with only one occurrence, once the first of the one quadrillion words is reached the CDF is going to increase really slowly (what is seen in the 1-grams). If there is only a 200 words corpus it is expected to shoot to 1 in the CDF by the 200th word. This appears to be why the smallest corpus (Talking) has the largest final slope because the amplitude is changing the most due to lack of infrequently used words.

*Table 1.* Frequency Rates of Growth for each Corpora

|  | **Rate 1** (Word$^{-1}$) | **Rate 2** (Word$^{-1}$) | **Final 100 Slope** (Word$^{-1}$) | **Final 150 Slope** (Word$^{-1}$) |
|---|---|---|---|---|
| 1-grams | 0.15625 | 0.015385 | 0.0005 | 0.000995 |
| Twitter | 0.10101 | 0.010684 | 0.0009 | 0.00145 |
| Chat | 0.114679 | 0.010526 | 0.00097 | 0.0015 |
| Talk | 0.125471 | 0.014493 | 0.001 | 0.001856 |

From this data I see that I need to correct for the discrepancy in corpus size because this changes the distribution and frequency of word usage, the things I want to compare. Using the first 200 words of each corpus as a representative population seems like a good assumption but do they represent what would be expected from a language corpus? To test this I used a power law distribution based on the zipfian distribution. This is what experts agree is a good approximation of language corpora since a long time ago (5). I fit the Usage frequency to the Power Law Equation:

$$\frac{1}{(a + Rank)^b}$$

Where Rank is the Rank of the word and a and b are fitting parameters. For this data a=6 and b=1.6.

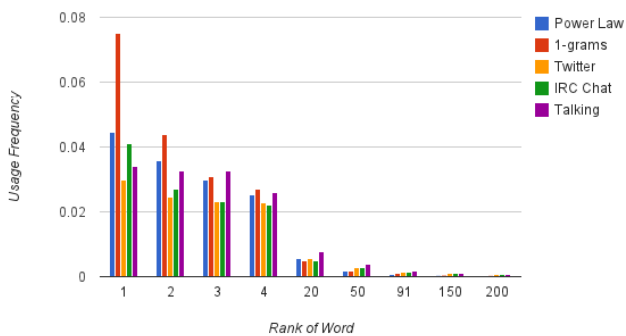

**Figure 2**. Comparing the usage frequency distribution of each corpus to a modeled Power Law distribution.

Though the usage frequency distributions are different, the population of the first 200 words follows a Power Law Distribution nicely. As these first 200 words comprise a majority of each corpus I believe it will not be unreasonable to use these populations as a representative of each corpus in further analysis. To make this work I calculated normalized frequencies based only on the population of the first 200 words. From these new populations similar trends as the full corpora are observed, such as the usage of the first 6 words in the 1-grams corpus is still a much larger part of the population, further verifying our model is a good representation (**Figure 1** and **Figure 3**).

How does word usage relate to communication? The ability of someone to construct diverse sentences depends not only on their vocabulary but on their word usage frequencies. An optimally diverse corpus would have an equal frequency of usage of each word. For example if we have a 3 word corpora and the usage frequency of each word is (0.60, 0.35, 0.05) the majority of sentences (95%) will only contain words 1(0.60) and 2(0.35) given a reasonable sentence length. If the frequencies are (0.33, 0.33, 0.33) each sentence of 3 words or more should theoretically contain all three words. A modality can be thought of as creating more diverse sentences by having a low root means square deviation(RMSD) in frequencies across the population compared to the optimal distribution (**Table 2**).

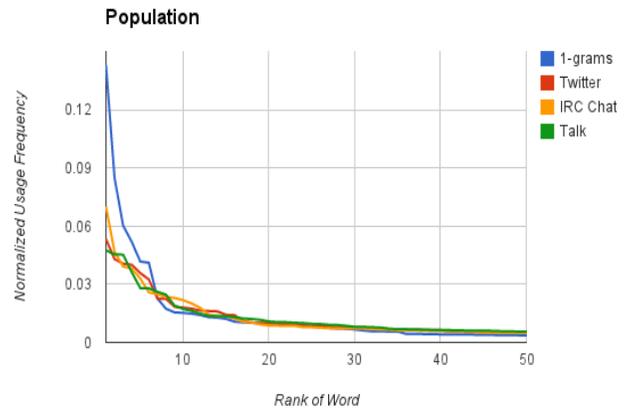

**Figure 3**. A graph of the normalized frequency of usage of each word contained in the 200 word population

**Table 2.** The root mean square deviation of population frequencies and for the lower and upper 50% of the population.

|  | RMS Dev. | Rank at 50% | RMS Dev. of Lower 50% (1-100) | RMS Dev. of Upper 50% (101-200) |
|---|---|---|---|---|
| **1-grams** | 0.014 | 12 | 0.0027 | 0.0392 |
| **Twitter** | 0.007 | 25 | 0.0018 | 0.0128 |
| **IRC Chat** | 0.008 | 24 | 0.0017 | 0.0152 |
| **Talk** | 0.007 | 26 | 0.0020 | 0.0120 |

In the populations, Google 1-grams seems to have the most deviant frequencies and so the least amount of sentence construction diversity. This is for the overall population and for the lower 50% (Words 1-100) and upper 50% (Words 101-200) (**Table 2**). I thought this might be because the 1-grams database is so large that maybe the First 200 words was composed entirely of pronouns or some such and so using the RMSD as a proxy for real sentence construction would be wrong but this is not the case (**Figure 5**). This is very strange as it means that authors cannot probabilistically construct as many different sentences as the other communication modalities. However, this is not an analysis of the exact sentences used in communication just a theoretical prediction based on the statistics of the corpora.

So in summary we see that Authors writing books, use fewer common words much much more often (these words are: *the, of, and, to, in, a*) and people who use Twitter, IRC

Chat or Talking appear to have more theoretical sentence construction diversity by having a more evenly distributed frequency of word usage.

*Matching words*

If we look at the # of matching words in the Top 200 populations we see that Twitter and Talk are the best matching with 142 out of 200 words matching or 71% (**Table 3**). While Twitter and 1-grams are the worst matching with only 115 or 57.5% words the same. Twitter and Chat seem very similar in composition of words, 70% matching.

*Table 3. Number of matching words between the Top 200 word populations.*

|  | 1-grams | Twitter | IRC Chat | Talk |
|---|---|---|---|---|
| **1-grams** |  | 115 | 119 | 128 |
| **Twitter** | 115 |  | 122 | 142 |
| **IRC Chat** | 119 | 122 |  | 140 |
| **Talk** | 128 | 142 | 140 |  |

*Copora Similarity and Confusion*

I created confusion matrices to determine not only if the words used in each corpus are similar but also if their frequency of usage is similar. The confusion matrices were created under the assumptions that if a word is randomly chosen from a random corpus and I am trying to determine which corpus the word came from based on frequency of usage alone not my *a priori* knowledge of where it came from, can I do it? And how successfully? An example would be if I randomly select a word from the Twitter corpus but based on our statistics I guess that it came from the 1-grams corpus then it is said that the word is "confused". It means I can't guess correctly based on my best guess method alone with no *a priori* knowledge.

I wanted to do this not only with one word but with multiple words to construct a sentence like analysis. So instead of looking for the probability of only one word I also looked for a combined probability of multiple words.

For the confusion matrix I ran 120,000 simulations each for 1 to 10 words which was ~30,000 per population of 200 words. For the simulations I randomly chose a population so the number is slightly above or below 30,000. After randomly choosing a population I randomly chose a word or group of words from it and then looked at each population to determine the usage frequency of the word. The population in which the word had the highest frequency was chosen. All corpora were compared simultaneously not head to head.

**Table 4**. Single word confusion matrix, frequency of confusion

| 1 Word | Twitter | Talk | 1-grams | IRC Chat |
|---|---|---|---|---|
| **Twitter** | 0.42 | 0.19 | 0.20 | 0.17 |
| **Talk** | 0.15 | 0.37 | 0.25 | 0.21 |
| **1-grams** | 0.11 | 0.13 | 0.57 | 0.17 |
| **IRC Chat** | 0.14 | 0.18 | 0.21 | 0.45 |

From the single word confusion matrix 1-grams has the least amount of confusion which agrees with our previous findings that the 1-grams population is different than the other populations and so matches itself more frequently. Populations tend to be confused when compared to 1-grams most likely due to the high frequency of some of the words. Talk is the most confused population being confused 63% of the total number of simulations. However, even Twitter has a high confusion percentage at 56% and IRC Chat at 53% (**Table 4**). When compared individually it appears that each population has the highest confusion with 1-grams. This is a statistical anomaly and probably does not relate to the true structure of word usage in populations because 1-grams has a disproportionately high frequency of usage for the first 5 to 10 ranked words(~2.5 – 5%) of the population as seen from our previous analysis. Looking at a multiword confusion matrix simulation of 4 words 1-grams becomes the least confused when comparing itself to others and now when comparing others to itself it becomes the least confused or next to least confused for each population. This effectively drowns out the effects of high frequency words. Twitter is confused most with Talk, Talk with IRC Chat and IRC Chat with Talk.

**Table 5**. Four word frequency of confusion from confusion matrix simulations

| 4 Words | Twitter | Talk | 1-grams | IRC Chat |
|---|---|---|---|---|
| **Twitter** | 0.79 | 0.10 | 0.05 | 0.06 |
| **Talk** | 0.08 | 0.71 | 0.09 | 0.12 |
| **1-grams** | 0.04 | 0.02 | 0.88 | 0.06 |
| **IRC Chat** | 0.04 | 0.07 | 0.05 | 0.83 |

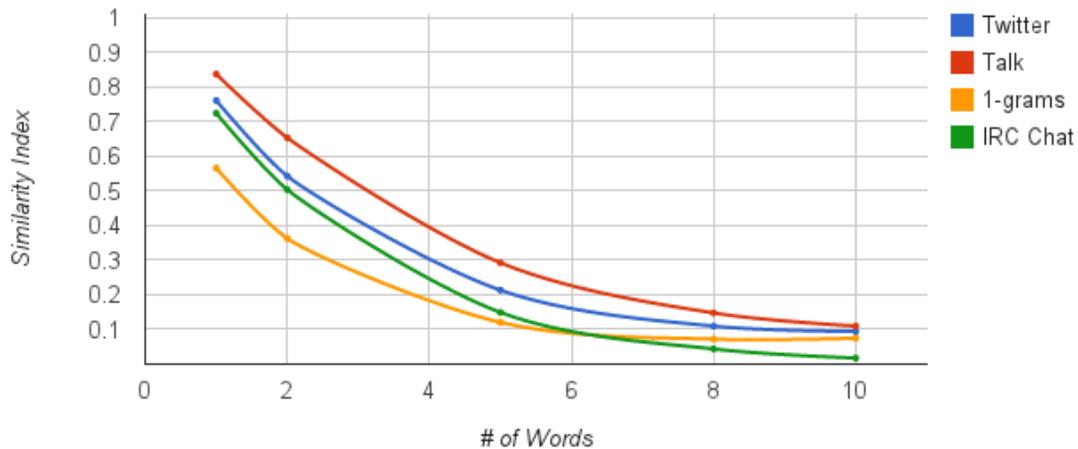

*Figure 4.* The Similarity Index of the 4 populations calculated from confusion matrices generated from 1, 2, 5, 8 and 10 words.

In order to compare the high order confusion matrices I created a single value to define one population's relation to all others.

I created a similarity index by this equation: $1 - \frac{\mu_d}{n_s}$

Where $\mu_d$ is the mean of the differences between the true positive and each false positive from the confusion matrix and $n_s$ is the number of simulations for the chosen population.

If all populations are identical they would be confused an equal proportion then we would expect a Similarity Index of 1 as $\mu_d$ would be close to 0. The further away from this equal distribution that a population is the larger $\mu_d$ and the smaller the Similarity Index value with a minimum near 0.

As the number of words used in a confusion simulation increase the similarity index should decrease because a population becomes more similar to itself and less similar to others as is seen in the graph. What the Similarity Index shows is that Talk is the most similar to all other populations regardless of number of words used in the simulated sentences (up to 10 words). Twitter and Chat start out similar but then diverge, while Talk and Twitter and eventually 1-grams all converge near 10 words. Chat and 1-grams also converge around 5 words but then diverge again by 10 words.

It is interesting that the overall structure (words, matching words, frequency of words) tends to be more similar for Twitter and Talk! Crazy! People tweet like they speak. The reason that the rates change over the course of the graph is that I am not randomly sampling unique words so a word can be chosen twice in a multiword simulation. Also, there is a higher probability of choosing a unique word, i.e. one that does not appear in any other corpora, the probability of which differs for each population. Initially, 1-grams are going to have a low value compared to the other populations on the Similarity Index because it has the least number of matching words. However, when the word count for a simulation increases the populations tend to be equally dissimilar so word frequency of these rare multi-word matching events will tend to dominate. Since 1-grams has the highest frequency in the most highly ranked words any probabilities calculated for a matching multi-word confusion simulation can be dominated by one of these words alone. For example, the word *the* in single word confusion simulations has a probability of randomly being chosen 1 in 200 times. This word will always be confused for 1-grams in this case because it has the highest frequency of usage. Whereas it has a probability of being chosen 1 in 20 times in the 10 word simulations. Because very few combinations of 10 words have a higher usage frequency than *the* alone it will dominate these simulations and is seen in the change in slope of Similarity Index decay of 1-grams compared to others.

Looking at the word class of the populations the number of nouns, adverbs and conjunctions are very similar (**Figure 5**). The main way to differentiate between these populations is by verb and adjective usage. In both verb and adjective usage 1-grams is an outlier compared to the other populations, using more adjectives and fewer verbs. The 1-grams and IRC Chat have a similar number of adjectives, pronouns and nouns.

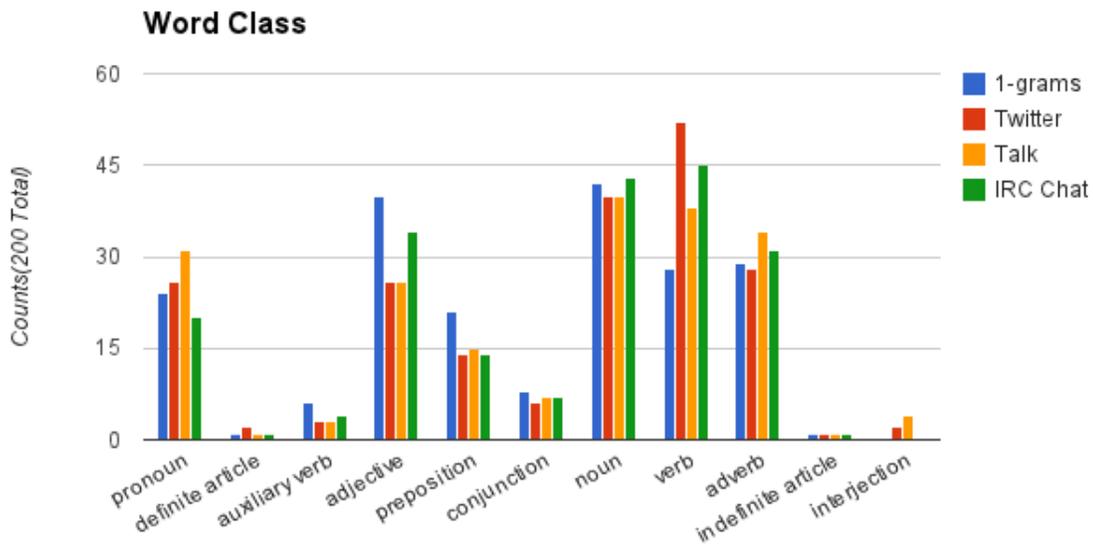

**Figure 5**. Word Class counts of the different populations

Overall, Twitter Talk and Chat populations tend to be similar in many aspects and the Google 1-grams population tends to be different. How well does this hold for a unique individual?

*Case Study*

In order to test out how pertinent this data is I wanted to test and see if a person actually uses different words and word usage frequencies when using these different modalities. I happened upon author Cory Doctorow and decided to use his corpora because he has open and free books available that I converted to 1-grams, a Twitter account and videos of him speaking and interacting on youtube.

All of the works of Mr. Doctorow match up with very similar distributions (**Fig. 6**). This was quite unexpected as compared to the amalgamized populations used above composed of many different authors. The 1-grams of Mr. Doctorow resemble Twitter and Talk more than they resemble the Google 1-grams. Again, the Google 1-grams distribution is completely different when compared to all the other distributions. Why is this? As seen with the Doctorow data it does not appear to be an intrinsic property of books. However, the books that are used of Mr. Doctorow are all fiction and so might bias the dataset.

I created Fiction and Nonfiction copora and performed word matching of the 200 word populations from Mr. Doctorow derived similarly as before. Again it was found that Talking is the most similar to the other communication modalities (**Table 4**). Talk has the most matching words between all the other corpora. Also, Twitter and Talk contain the most matching words between populations. However, it is interesting to see that both Fiction and Nonfiction works by Mr. Doctorow are comparable in their similarity to the Talking vocabulary and also similar in their word matching of Twitter (**Table 4**).

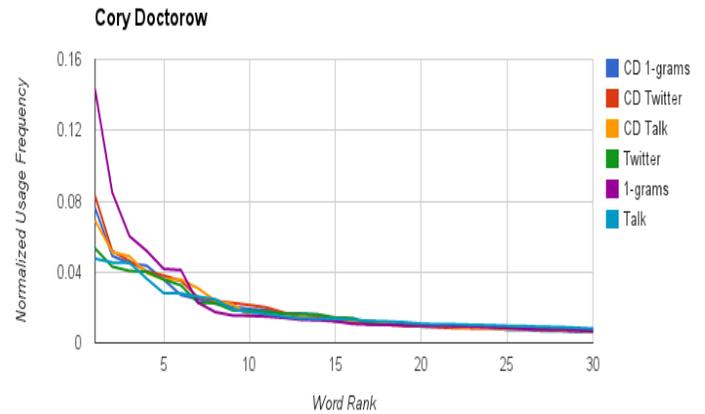

**Figure 6.** Population of the first 200 words graphed as normalized usage frequency for works of Cory Doctorow (CD). Twitter, 1-grams and Talk are the 200 word populations derived earlier. Only the first 30 words are plotted to emphasize the initial distributions.

**Table 4.** Matching words from Cory Doctorow populations only

|  | Fiction | Nonfiction | Twitter | Talk |
| --- | --- | --- | --- | --- |
| **Fiction** |  | 125 | 122 | 139 |
| **Nonfiction** | 125 |  | 126 | 140 |
| **Twitter** | 122 | 126 |  | 160 |
| **Talk** | 139 | 140 | 160 |  |

To look for differences between the Fiction and Nonfiction populations of Mr. Doctorow I analyzed the word class of the nonmatching words and found that the usage of verbs and nouns between the two was the most different (**Table 5**). In the original populations, the 1-grams population has fewer verbs and more nouns compared to the others, similar to what is found between these two populations.

**Table 5**. Word Class of nonmatching words between the Cory Doctorow fiction and nonfiction corpora

|  | Noun | Pronoun | Verb | Adverb | Adjective |
|---|---|---|---|---|---|
| **Fiction** | 20 | 9 | 29 | 8 | 9 |
| **Nonfiction** | 44 | 4 | 15 | 4 | 7 |

*Fiction and Nonfiction*

Based on results from Mr. Doctorow's corpora I decided to make a collection of amalgamized Fiction and Nonfiction works to see how different they are and compare it to the single author populations. The Fiction corpus is composed of 5 books by 2 different authors and the Nonfiction books corpus is composed of 5 books from 5 different authors. By comparing the works of Mr. Doctorow to the multi-author populations we see that his Fiction tends to be easily identifiable as fiction with a 77% word match, while his Nonfiction is more unique (**Table 6**) with some of the lowest word matching counts seen. The reason Mr. Doctorow's nonfiction works might be different than the 1-grams or Nonfiction populations could be a result of written books being the primary composition of those populations and Mr. Doctorow's Nonfiction is composed mostly of news writing, exposition and opinion.

**Table 6.** Works of Cory Doctorow compared to Non-Cory Doctorow populations.

|  | Fiction | Nonfiction | 1-grams | Twitter | Talk | IRC Chat |
|---|---|---|---|---|---|---|
| **CD Fiction** | 154 | 113 | 123 | 135 | 144 | 130 |
| **CD Nonfiction** | 111 | 110 | 119 | 111 | 114 | 116 |

The frequency of the first 200 words of the amalgamized Fiction and Nonfiction populations show that the distribution of the Nonfiction population matches up very closely to the 1-grams population in the first 5 deviated words, a characteristic of the 1-grams population, and the Fiction does not (**Fig. 7**). Looking at the actual number of matching words in the population it agrees with the distribution plot showing that the Nonfiction corpus matches poorly with Twitter, Chat and Talk while the Fiction corpus matches well with all the corpora (**Table 7**). Mr. Doctorow's Nonfiction work did not match well with the distribution of words in the 1-grams population. This further suggests that the form of nonfiction collected from Mr. Doctorow is a different form of communication as compared to nonfiction book writing.

**Table 7.** Number of Matching Words in the population of the first 200 words

|  | 1-grams | Twitter | Talk | IRC Chat | Fiction |
|---|---|---|---|---|---|
| **Fiction** | 130 | 133 | 146 | 133 | ■ |
| **Nonfiction** | 143 | 107 | 115 | 107 | 123 |

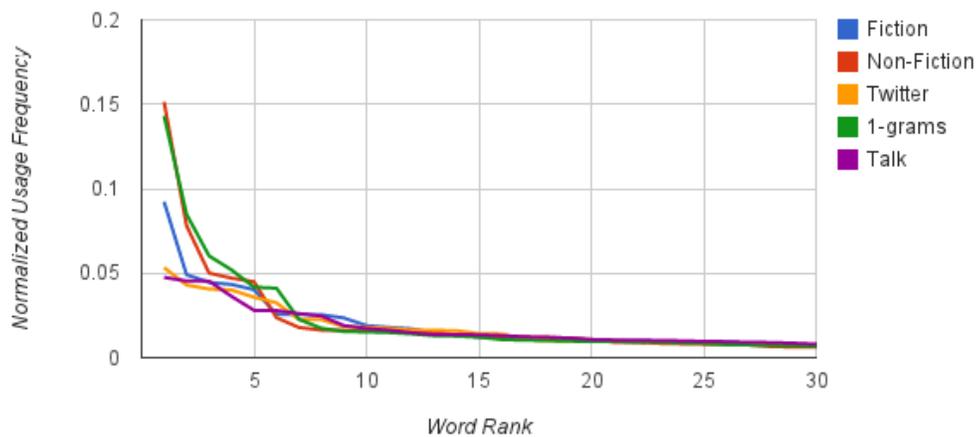

**Figure 7**. A graph of the word usage frequency of all populations including Fiction and Nonfiction.

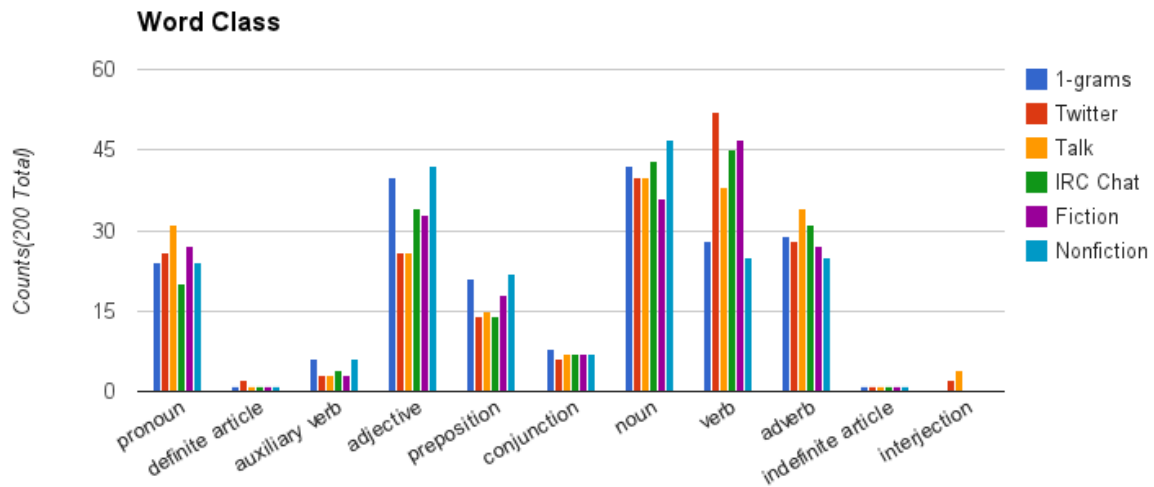

**Figure 8**. Word class of all amalgamized populations including fiction and nonfiction

The word class data with Fiction and Nonfiction works suggest further that the Google 1-grams population is composed mostly of nonfiction works (**Figure 8**). The Fiction population matches closely to IRC Chat but also has a similar pattern as Talk and Twitter. The Nonfiction seems to be very similar to 1-grams in word class usage. Combined this data suggestively identifies that the 1-grams corpus from the Google Books project is composed mostly of Nonfiction works and so skews the distribution of words and words usage frequencies as compared to communication using other modern language modalities. Overall, though our lives are composed of nonfiction actions and elements, works of fiction tend to more closely resemble Talking and modern communication modalities, such as Twitter and IRC Chat. It begs the question, Could nonfiction books written with fiction attributes be considered more entertaining?

Based on this analysis the data suggests that it is possible to discriminate the basic composition of a book, i.e. whether it is fiction or nonfiction based on different properties of vocabulary alone without looking at complex sentence structure or other factors. It also brings up a few interesting questions: Can this possibly be extended further to classify books by genre and can these techniques be used by computers to automatically annotate and categorize books like is done for DNA sequences? Can these analyses be used to validate that books written are actually nonfiction works?

**Acknowledgements**

I would like to thank Melissa Runfeldt and Cory Doctorow for comments on the manuscript.